\pdfoutput=1
\documentclass[11pt]{article}

\usepackage[final]{acl}

\usepackage{times}
\usepackage{latexsym}

\usepackage[T1]{fontenc}
\usepackage[utf8]{inputenc}

\usepackage{microtype}

\usepackage{inconsolata}

\usepackage{booktabs}       % professional-quality tables
\usepackage{amsfonts}       % blackboard math symbols
\usepackage{nicefrac}       % compact symbols for 
\usepackage{graphicx}      % microtypography
\usepackage{xcolor}         % colors
\usepackage{graphicx}
\usepackage{multirow}
\usepackage{subfig}
\usepackage{lineno}
\usepackage{pifont} 
\usepackage{hyperref}
\usepackage{color}
\usepackage{colortbl}
\usepackage{graphicx}
\usepackage{booktabs}
\usepackage{multirow}
\newcommand{\checkmarkgreen}{\textcolor{green}{\checkmark}}%
\title{Llama SLayer 8B: Shallow Layers Hold the Key to Knowledge Injection}

\author{
  \textbf{Tianxiang Chen\textsuperscript{1,2,3,4}}\thanks{Work done during an internship at Alibaba Cloud.},
  \textbf{Zhentao Tan\textsuperscript{1,2,3,4}},
  \textbf{Tao Gong\textsuperscript{1,2,3}}\thanks{Tao Gong is the corresponding author.},
  \textbf{Yue Wu\textsuperscript{4}},
  \textbf{Qi Chu\textsuperscript{1,2,3}},
  \textbf{Bin Liu\textsuperscript{1,2,3}}\\
  \textbf{Jieping Ye\textsuperscript{4}},
  \textbf{Nenghai Yu \textsuperscript{1,2,3}}
\\
  \textsuperscript{1}School of Cyber Science and Technology, University of Science and Technology of China\\
  \textsuperscript{2}Anhui Province Key Laboratory of Digital Security\\
  \textsuperscript{3}CAS Key Laboratory of Electromagnetic Space Information\\
  \textsuperscript{4}Alibaba Cloud
\\
  \small{
    \textbf{Correspondence:} \href{mailto:email@domain}{tgong@ustc.edu.cn}
  }
}

\begin{document}
\maketitle
\begin{abstract}
As a manner to augment pre-trained large language models (LLM), knowledge injection is critical to develop vertical domain large models and has been widely studied. Although most current approaches, including parameter-efficient fine-tuning (PEFT) and block expansion methods, uniformly apply knowledge across all LLM layers, it raises the question: are all layers equally crucial for knowledge injection? We begin by evaluating the importance of each layer in finding the optimal layer range for knowledge injection. Intuitively, the more important layers should play a more critical role in knowledge injection and deserve a denser injection. We observe performance dips in question-answering benchmarks after the removal or expansion of the shallow layers, and the degradation shrinks as the layer gets deeper, indicating that the shallow layers hold the key to knowledge injection. This insight leads us to propose the S strategy, a post-pretraining strategy of selectively enhancing shallow layers while pruning the less effective deep ones. Based on this strategy, we introduce Llama Slayer-8B and Llama Slayer-8B-Instruct. We experimented on the corpus of code $\&$ math and demonstrated the effectiveness of our strategy. Further experiments across different LLM, Mistral-7B, and a legal corpus confirmed the general applicability of the approach, underscoring its wide-ranging efficacy.
Our code is available at: \href{https://github.com/txchen-USTC/Llama-Slayer}{https://github.com/txchen-USTC/Llama-Slayer}.
\end{abstract}

\section{Introduction}
Large Language Models (LLMs)
have significantly revolutionized the natural language processing area, showcasing unparalleled abilities across various tasks \cite{achiam2023gpt}. Despite their versatility, LLMs
exhibit limitations in specialized areas such as mathematics, programming, etc., which hinder the potential of wide-ranging applications. To address these gaps, existing work \cite{liu2023llm360,wang2023huatuo} has sought to enhance the diverse skills of pre-trained LLMs through customized data strategies. However, they require extensive computational efforts and massive data volumes, challenging the widespread accessibility of LLM research. Furthermore, while Parameter-Efficient Fine-Tuning (PEFT) techniques offer a reduction in training requirements, their effectiveness tends to diminish \cite{biderman2024lora, wu2024llama} compared to traditional fine-tuning methods, especially as the size of the model and the dataset grows. 

\begin{figure}
    \centering \includegraphics[width=0.50\textwidth,height=0.23\textwidth]{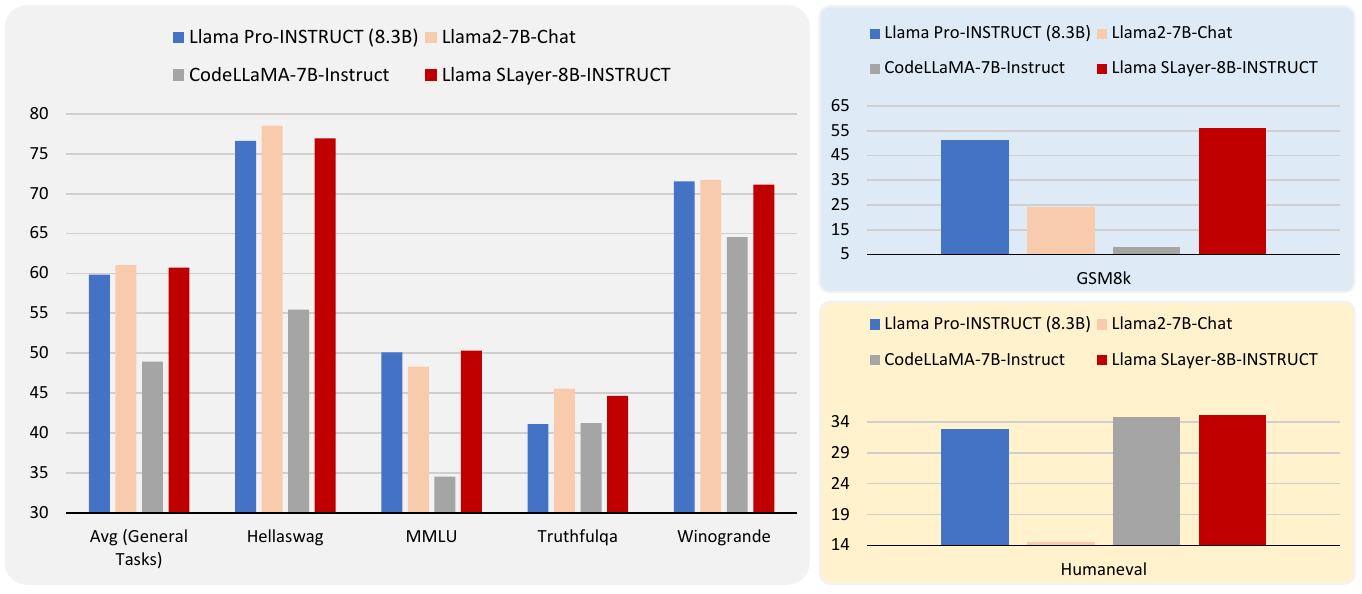}
    \caption{Llama SLayer-8B-INSTRUCT achieves state-of-the-art performance across several tasks, spanning from general language tasks to specific domain tasks (programming and mathematics), outperforming its LLaMA series predecessors.}\label{fig1}%These plots illustrate abrupt performance shifts in performance for the accuracy of QA tasks.
\end{figure}

% Subsequently, another line of research, known as model merging methods \cite{kim2023solar,wu2024llama,akiba2024evolutionary}, has emerged. These methods refine pre-trained models with domain-specific datasets in the post-pretraining phase. These strategies tend to involve more trainable parameters than PEFT methods, yet not as many as those used in full fine-tuning processes. These approaches enhance the performance of domain-specific downstream tasks while mitigating computational expenses.

Subsequently, another line of research emerged, focusing on methods such as model merging \cite{akiba2024evolutionary} and model expansion \cite{wu2024llama,choi2024life,kim2023solar}. Model merging methods strive to synthesize a multifaceted model that amalgamates insights from various pre-trained domain-specific LLMs, potentially crafting a model adept at addressing a multitude of tasks concurrently. However, the process of training multiple domain-specific LLMs is resource-intensive. On the other hand, model expansion methods, exemplified by Llama Pro, seek to refine pre-trained models for domain-specific applications in the post-pretraining phase by only fine-tuning the expanded layers. Therefore, it can employ significantly fewer trainable parameters than full model fine-tuning.

%Nonetheless, present model expansion methods generally treat each part of LLMs equally, though different layers may exhibit varying sensitivity to the incorporated knowledge. This lack of differentiation can result in less-than-ideal knowledge injection results. An intuitive idea is to inject knowledge into the most important layers so that the LLM can more sufficiently leverage the new knowledge without the overhead of redundant adjustments across all layers. To this end, we select three different metrics to evaluate the importance of each layer to find which part of LLM is more important to knowledge injection, and our experiment results are shown in Fig.~\ref{charts}. To ensure generality, we conducted evaluations on two types of LLMs: Llama2-7B and Mistral-7B. 

%Our evaluation involved measuring the angular distance between feature inputs of adjacent blocks throughout all layers of LLMs and revealed that both the shallow and the final layers exhibit higher angular distances. This finding indicates the paramount importance of these layers, as they undergo more significant modifications in response to new data exposure. We also tested the accuracy drop of the two LLMs dropping one arbitrary layer or inserting one more arbitrary layer on two general QA benchmarks.

However, present model expansion methods generally treat each part of LLMs equally, although different layers may exhibit varying sensitivity to incorporated knowledge. This lack of differentiation can result in less-than-ideal knowledge injection results. An intuitive idea is to inject knowledge into the most important layers so that the LLM can more sufficiently leverage the new knowledge without the overhead of redundant adjustments across all layers. To this end, we select three different metrics to evaluate the importance of each layer to find which part of the LLM is more important to knowledge injection. Our findings suggest that the shallow layers are more important compared to the last few layers, as the drop in precision - whether through the removal or addition of the last few layers - is markedly less significant than the drops induced by manipulating the shallow layers. Drawing on this insight, we propose S strategy, a novel strategy to knowledge injection that concentrates on enriching the shallow layers while deleting the least important deepest layers. Based on the proposed strategy, we further introduce Llama SLayer-8B and Llama SLayer-8B-INSTRUCT, versatile LLMs that excelling in programming, and mathematics and general language tasks. Figure~\ref{charts} displays the superiority of Llama SLayer-8B-INSTRUCT.

The main contributions of this paper can be summarized to three aspects:
\begin{itemize}
    \item We propose a novel post-pretraining strategy
for LLMs, namely S strategy, that focuses knowledge injection to the important layers while pruning the ineffective layers.
    \item Based on our S strategy, we introduce Llama SLayer-8B and Llama SLayer-8B-INSTRUCT, versatile LLMs that excelling in programming, and mathematics and general language tasks.
    \item We benchmark the family of Llama SLayer
on extensive datasets, demonstrating its exceptional performance and significant promise for diverse and complex applications.
\end{itemize}

% \begin{figure}
%     \centering \includegraphics[width=0.49\textwidth,height=0.64\textwidth]{charts.pdf}
%     \caption{(a) Angular distance between feature inputs of adjacent blocks, vs. the layer number. The features are obtained by making inference on the MMLU question-answering benchmark. (b) Accuracy of Llama2-7B and Mistral 7B without one layer on two general QA benchmark; here, the dashed blue and green lines indicate the accuracy of the original baseline 7B models on two general QA benchmarks. (c) Accuracy of Llama2-7B and Mistral 7B with one more expanded block on the same two QA benchmarks. The initialization of the expanded blocks include identity copy and averaging the adjacent block weights.}\label{charts}%These plots illustrate abrupt performance shifts in performance for the accuracy of QA tasks.
% \end{figure}

\section{Related Works} 
Here we introduce four prevalent types of methods for injecting domain-specific knowledge, including full fine-tuning, parameter-efficient fine-tuning, model merging, and model expansion.
\subsection{Full Fine-tuning Methods} 
% Full fine-tuning of pretrained language models (PLM) requires training all parameters on a specific downstream task using domain-specific knowledge. These models are initially trained on large datasets with unsupervised learning objectives like language modeling or masked language modeling, to learn general language representations \cite{touvron2023llama,liu2019roberta}. However, when these generically trained PLMs are steered toward specific downstream tasks, their performance might fall short due to the absence of specialized domain knowledge \cite{xu2023improving,dabre2019exploiting}. Full fine-tuning provides an effective solution to address this limitation. For example, HuaTuo \cite{wang2023huatuo} is a Chinese biomedical LLM tuned from LLaMA-7B tuned with Chinese Medical Knowledge. CodeLLama \cite{roziere2023code} and Code-Qwen \cite{li2023starcoder} are LLMs for code generation and infilling tuned from Llama and Qwen. However, full fine-tuning necessitates substantial computational
% resources and labeled data, as the model is trained from scratch for the specific target task. Moreover, as PLMs grow in size and with the advent of LLMs containing billions of parameters, full fine-tuning places even greater demands on computational resources. Furthermore, full fine-tuning may give rise to over-fitting when the task-specific dataset is small or when the PLMs are already well-suited to the target task \cite{pfeiffer2020adapterfusion}.

Full fine-tuning of Pretrained Language Models (PLMs) involves retraining all parameters for a particular task with domain-specific knowledge \cite{touvron2023llama,liu2019roberta}. Initially trained on vast unsupervised datasets to learn broad language representations, these PLMs may underperform on specialized tasks due to lack of domain-specific expertise \cite{xu2023improving,dabre2019exploiting}. Full fine-tuning addresses this by adapting models such as HuaTuo \cite{wang2023huatuo}, a Chinese biomedical LLM based on LLaMA-7B, and programming-focused LLMs such as CodeLLama \cite{roziere2023code} and Code-Qwen \cite{li2023starcoder}, for targeted applications. Despite its effectiveness, this approach requires extensive computational resources and substantial labeled data, which pose challenges like overfitting on small task-specific datasets, particularly as PLMs increase in size and complexity \cite{pfeiffer2020adapterfusion}.

\subsection{Parameter-Efficient Fine-Tuning Methods} 
% To alleviate the tremendous demands for computational
% resources, PEFT methods can selectively update or modify specific parts of the PLMs while still achieving performance comparable to full fine-tuning \cite{zaken2021bitfit}. Inspired by Intrinsic SAID \cite{aghajanyan2020intrinsic}, LoRA \cite{hu2021lora} introduces two trainable low-rank matrices for weight update. During training, the weights of PLM
% are frozen, and only the low-rank matrices of LoRA are trained. AdaLoRA \cite{zhang2023adaptive} extends LoRA by dynamically adjusting the rank of matrices to control the allocation. Delta-LoRA \cite{zi2023delta} updates the pre-trained weight W and two low-rank matrices $W_{down}$ and $W_{up}$ while using the same memory as LoRA. Adapter-based Fine-tuning \cite{houlsby2019parameter,lei2024conditional}, in which the adapter module is incorporated into the transformer, allowing for fine-tuning without modifying
% the pre-trained parameters. However, once the scale of model and datasets becomes huge, PEFT methods substantially under-perform full fine-tuning \cite{biderman2024lora, wu2024llama}. 

To reduce computational demands, Parameter-Efficient Fine-Tuning (PEFT) techniques modify only trivial parts of PLMs, maintaining performance comparable to full fine-tuning. LoRA \cite{hu2021lora} uses trainable low-rank matrices to update weights with the original weights of the PLM remaining unchanged, while AdaLoRA \cite{zhang2023adaptive} adjusts the rank of these matrices for optimized performance. %Delta-LoRA modifies the approach by updating both the pre-trained weights and low-rank matrices with efficient memory usage. 
Adapter-based Fine-tuning \cite{houlsby2019parameter,lei2024conditional} introduces adapters into the transformer architecture, allowing fine-tuning with minimal alteration to pre-trained parameters. However, as the size of models and datasets increases significantly, PEFT methods tend to fall behind in performance compared to full fine-tuning approaches \cite{biderman2024lora, wu2024llama}. %This is largely because the number of trainable parameters in PEFT methods is insufficient to encapsulate the vast amount of specific knowledge required at larger scales. 

\subsection{Model Merging Method} 

Model merging methods aim to create a comprehensive model by integrating knowledge from several pre-trained domain-specific LLMs. Task Arithmetic \cite{yadav2024ties} construct task vectors by differentiating pre-trained and fine-tuned model weights, allowing for model behavior adjustments through arithmetic operations. DARE \cite{yu2023language} further refines this by focusing on and enhancing the critical disparities between models. Evolutionary algorithms proposed by Takuya et al. \cite{akiba2024evolutionary} streamline the merging process without necessitating fine-tuning, although this method's reliance on multiple pre-trained models and the subjective nature of the merging strategy may complicate its broad use. However, generating multiple domain-specific LLMs still requires substantial computational resources.

\subsection{Model Expansion Method} 

%Despite PEFT methods can achieve performances comparable to full fine-tuning only in cases where the model and the knowledge corpus is of relatively small scale. Once the model and dataset scales increase, PEFT methods gradually lags far behind full fine-tuning \cite{biderman2024lora, wu2024llama} since the trainable parameters are far from enough to hold so much specific knowledge. To narrow the gap, model merging methods are proposed. These methods often feature significantly higher trainable parameters than PEFT methods, but still less than full-tuning and can bring amazing performances. 

% Although Parameter-Efficient Fine-Tuning (PEFT) methods can attain results comparable to full fine-tuning, their effectiveness diminishes as the size of the model and dataset grows \cite{biderman2024lora, wu2024llama}-PEFT approaches start to significantly lag behind full fine-tuning as model and data scales increase. This is largely because the number of trainable parameters in PEFT methods is insufficient to encapsulate the vast amount of specific knowledge required at larger scales. 

To reach a better trade-off between computational resources and domain-specific performance, the model expansion method has been introduced. These techniques typically incorporate a more trainable parameters than PEFT methods, albeit significantly fewer than what is employed in full fine-tuning, and have been shown to yield impressive results. SOLAR 10.7B \cite{kim2023solar} features an innovative approach that involves merging the initial 24 layers with the final 24 layers of the same model in depth as its continuous pre-training strategy. However, its superior performance comes at the cost of training all 48 layers after expansion. Llama Pro \cite{wu2024llama} adopts a method of evenly distributing expansion blocks across all layers of the model. These expansion blocks are initialized by duplicating the weights of the preceding block and zeroing out specific weights to guarantee the same initial output as the original base model. During the subsequent phase of continual pre-training, only these expanded blocks get trained. 

However, present model expansion methods lack exploration on which part of layers is more suitable for merging, since different layers may not be equally sensitive to the injected knowledge. To explore which parts of LLMs are pivotal for knowledge injection, we evaluated different LLMs based on layer importance and discovered that shallow layers wield greater importance than deep layers. Based on this, we propose a novel knowledge injection strategy, namely the S strategy, that targets the knowledge injection to the important layers via block expansion and dispenses with the ineffective last few layers. We implement this strategy and propose Llama Slayer 8B to validate its effectiveness-enabling LLMs to better specialize in specific tasks while preserving general abilities.

\section{Method}
\label{headings}

%According to our extensive experiments and analysis, the conclusion is that the shallow layers are generally more important than deep layers for knowledge injection. Building on this insight, we enhance these crucial layers through block expansion, while eliminating a select few layers identified as less important. Only the parameters of the expanded layers get updated for the downstream task, leaving the rest frozen. 

%\subsection{Importance-Based Layer Selection $\&$ Deletion}

\subsection{Evaluation Metrics of Layer Importance}

%To locate which part of the layers is more important in terms of knowledge injection, we choose three different metrics. We aim to converge on a unified conclusion from these metrics, thereby identifying the range of layers that are of utmost importance. This will lay the groundwork for our S strategy.

%\subsubsection{Layer Importance Derived from Angular Distance}

\subsubsection{Angular Distance} 
%To ascertain which parts of the layers are more important to knowledge injection in the continual pre-training phase, we decide to evaluate from the feature transition aspect. To this end, we have adapted an angular distance (AD) metric to evaluate the significance of each block. The angular distance $\mathcal A_{i,i+1}$ between the input features of block $i$ and block $i+1$ is calculated as follow:

We first try to evaluate layer importance from the feature transition aspect. To this end, we adopt the angular distance (AD) metric to evaluate the significance of each block. The angular distance $\mathcal A_{i,i+1}$ between the input features of block $i$ and block $i+1$ is calculated as follows:

\begin{equation}\label{eq:angular}
\mathcal A_{i,i+1} =\frac{1}{\pi} arccos(\frac{x_{i}^{T}x_{i+1}}{||x_{i}||||x_{i+1}||})\\
\end{equation}
where $|| . ||$ denotes the $L_2$-norm. This metric helps identify blocks where significant data processing shifts occur when exposed to new data, indicating their pivotal role in adapting to new knowledge. We calculate AD using the MMLU test benchmark to get a low-fluctuation estimate. Higher AD denotes higher difference between between inputs and outputs of each block, therefore the areas with higher angular distance are earmarked for modifications, such as layer expansion, to augment the model's adaptability and improve its performance on the specific domain dataset. %Experiments indicate that the initial half of all layers, along with the final layer, hold greater importance.

%This distance needs to be accumulated over a number of examples that are large enough to get a low-fluctuation estimate.  

\begin{figure}
    \centering \includegraphics[width=0.38\textwidth,height=0.53\textwidth]{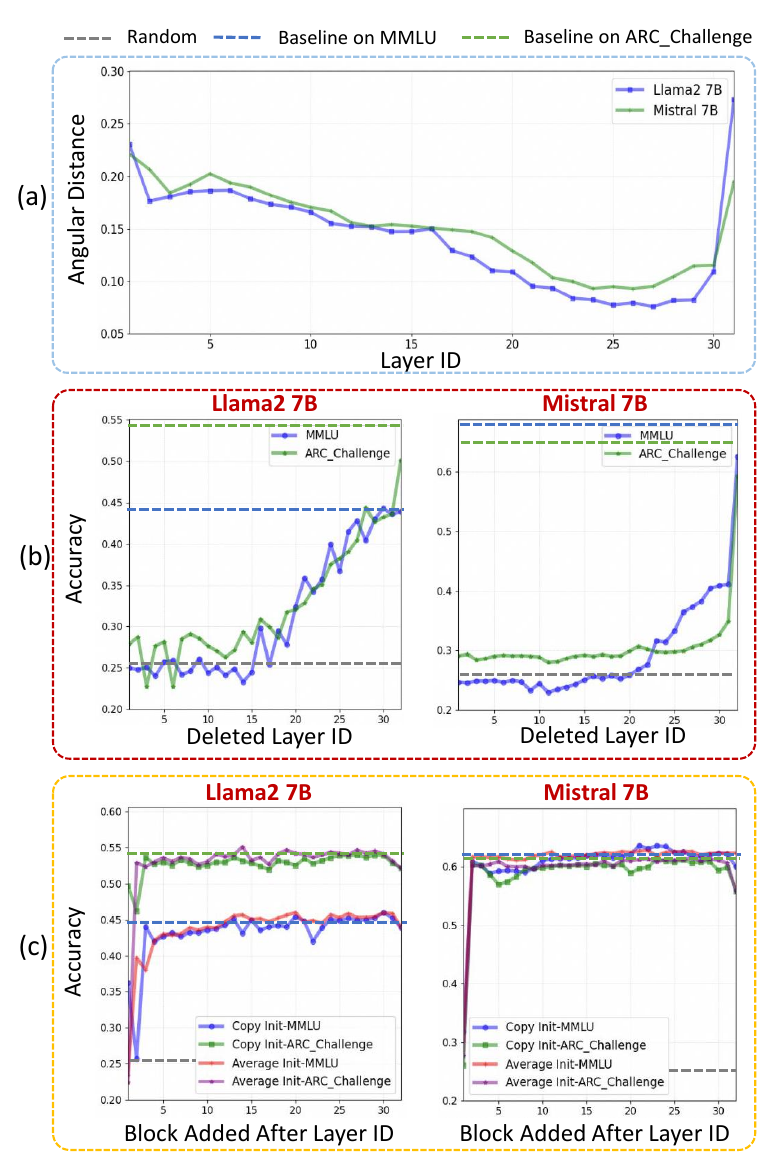}
    \caption{(a) Angular distance between inputs and outputs of each block vs. the layer number; (b) Accuracy of Llama2-7B and Mistral 7B dropping one layer on two general QA benchmarks; (c) Accuracy of Llama2-7B and Mistral 7B with one more inserted block on the same two QA benchmarks. The initialization of the expanded blocks includes identity copy and averaging the adjacent block weights.}\label{charts}%These plots illustrate abrupt performance shifts in performance for the accuracy of QA tasks.
\end{figure}

\subsubsection{Performance Drop after Layer Removal}
We can also locate the important layer areas by comparing the overall model performance drop on general question-answering benchmarks when removing different layers. We consider the layers that exhibit the most significant drop in performance once they are deleted as the most critical. Through experiments on Llama2 7B and Mistral 7B on two general QA benchmarks (MMLU \cite{hendrycks2020measuring} and ARC \cite{clark2018think}), we find the shallow layers as the important ones, so we locate the first half of all layers as the core area for knowledge injection. The choice of these two benchmarks stems from observing that the LLM after layer deletion or insertion has lost logic inference ability (evidenced by zero accuracy on GSM8k \cite{cobbe2021training} and Humaneval \cite{chen2021evaluating}), so we select the two multi-choice QA general benchmarks to measure performance drops.

Our intuition for layer removal comes from thinking about the representations as slowly changing
the function of layer index. In particular, the layer-to-layer evolution of representations for a transformer
is given by a residual iteration equation
\begin{equation}\label{eq:res}
x_{i+1} = x_{i}+F(x_{i},\theta_{i})\\
\end{equation}
where $x_{i}$ and $\theta_{i}$ are the input and parameter vectors for layer $i$, respectively, and $F(x_{i},\theta_{i})$ describes the transformation of one multi-head self-attention and MLP layer block. If we remove layer $i$, then we must now connect the old input to that layer, $x_{i-1}$, into the block function of layer $i+1$ as
\begin{equation}\label{eq:remove}
x_{i+1} = x_{i-1}+F(x_{i-1},\theta_{i-1})\\
\end{equation}
Comparing Equation ~\ref{eq:res} with Equation ~\ref{eq:remove} we can find a mismatch between the original input and new input, which should be very damaging for the network and cause a performance drop.

\subsubsection{Performance Drop after Layer Insertion}

Layer insertion is the reverse process of layer removal.
In particular, the layers that suffer the most pronounced performance drop when testing on MMLU \cite{hendrycks2020measuring} and ARC \cite{clark2018think} when inserting a new block after it are deemed the most crucial.

%In particular, the layer importance can also be determined by assessing the impact of expanding each layer on the overall performance in general question-answering benchmarks. Those layers that exhibit the most pronounced drop in performance upon expansion are deemed the most crucial.

Two different initialization methods can be employed for the insertion. The first is the identity copy of the weights from the preceding block. If we expand a block between layer $i$ and $i+1$, the layer-to-layer evolution of representations from layer $i$ to $i+1$, initiated by identity copy, can be depicted as
\begin{equation}\label{eq:add}
x_{i+1} = x_{i}+F(x_{i},\theta_{i})+F(x_{i}+F(x_{i},\theta_{i}),\theta_{i})\\
\end{equation}
In addition, we also employs weight averaging as the second method for the inserted block by averaging the weights of the adjacent two blocks. The reason for this initialization is that we think identity copy may be not smooth enough and we hope to start the later continual pre-training from a more smooth initialization state. In this way, the representation evolution from layer $i$ to $i+1$ is

\begin{equation}\label{eq:add}
x_{i+1} = x_{i}+F(x_{i},\theta_{i})+F(x_{i}+F(x_{i},\theta_{i}),\frac{\theta_{i}+\theta_{i+1}}{2})\\
\end{equation}

%Our observations reveal that the performance drop in the first half of the layers is less pronounced compared to the latter half, indicating that the first half shallow layers are of greater importance. Moreover, the weight averaging method of initialization has been found to result in a smaller performance decline than the identity copy approach, thus motivating us to favor weight averaging as the block expansion initialization method going forward.

%从持续学习的角度（不要直接pretraining吧，因为现在有些文章也提到在sft注入知识）划分几类，一个是全参数微调，可以引用几个领域大模型的例子，一个是高校参数微调，一个是模型扩展（llamapro和拼羊驼的那篇）

\subsection{Evaluation Analysis}

%Observing the angular distances across the layers, we note that the first half and the final layer exhibit higher angular distances, signaling their higher importance. The performance impact of removing layers showcases a more noticeable drop in the initial half, further emphasizing the critical role of the initial shallow layers. Similarly, by observing the performance drop after layer insertion, we find layer insertion mirrors this trend of performance drop, which again presents the importance of the shallow layers. Additionally, initializing with the weight averaging method has demonstrated less impact on performance compared to the identity copy method, thereby guiding our preference towards adopting weight averaging for initializing expanded blocks in future endeavors.

We illustrate the layer importance evaluation results in Figure~\ref{charts}. Observing the angular distances across the layers (Figure~\ref{charts} (a)), we note that the first half and the final layer exhibit higher angular distances. The performance impact of removing the layers (Figure~\ref{charts} (b)) shows a more noticeable drop in the initial half than in the last half of the layers. Similarly, observing the performance drop after layer insertion (Figure~\ref{charts} (c)), we find that layer insertion mirrors this trend of performance drop. Additionally, initializing with the weight-averaging method has demonstrated less impact on performance compared to the identity copy method. %thereby guiding our preference towards adopting weight averaging for initializing expanded blocks in future endeavors.

According to the above observations, we conclude that the shallow layers play a more crucial role in knowledge injection compared to the deeper layers. This is evidenced by (1) the angular distances are generally higher in shallow layers, indicating a more significant shift in features in these areas, and (2) the elimination or inserting of shallow layers poses more severe impacts on the overall performance of LLM in two QA benchmarks, underscoring the pivotal role of shallow layers and the relative ineffectiveness of deep layers. Given the ineffectiveness of the deepest layers, we opt to prune them prior to post-pretraining to assess the impact on final performance. We also observe a more modest decline in performance when layers inserted via weight averaging initialization are compared to those initialized with an identity copy, suggesting that weight averaging introduces additional coherence to the expanded LLM. This prompts further exploration into the potential benefits of using weight averaging as an initialization strategy for post-pretraining. %Building on these insight, we enhance these crucial layers through block expansion, while eliminating a select few layers identified as less important. Only the parameters of the expanded layers get updated for the downstream task, leaving the rest frozen. 

%Our comprehensive experiments lead to the conclusion that shallow layers play a more crucial role in knowledge injection compared to deeper layers. This is evidenced by (1) the observation that angular distances are generally higher in shallow layers, indicating a more significant shift in features in these areas, and (2) the fact that modifications to shallow layers—either through deletion or insertion—have a more profound impact on the overall performance of Large Language Models (LLMs) across two QA benchmarks, underscoring the essential function of shallow layers and the relative ineffectiveness of the deepest layers. Given the lesser effectiveness of the deepest layers, we explored pruning these layers before further pre-training to assess any potential impact on final performance. Additionally, we observed a more modest decline in performance when layers inserted via a weight averaging method compared to those initialized with an identity copy, suggesting that weight averaging introduces additional coherence to the expanded LLM. This finding prompts further investigation into the potential benefits of using weight averaging as an initialization strategy for post-pretraining enhancements.

\begin{figure}
    \centering \includegraphics[width=0.41\textwidth,height=0.51\textwidth]{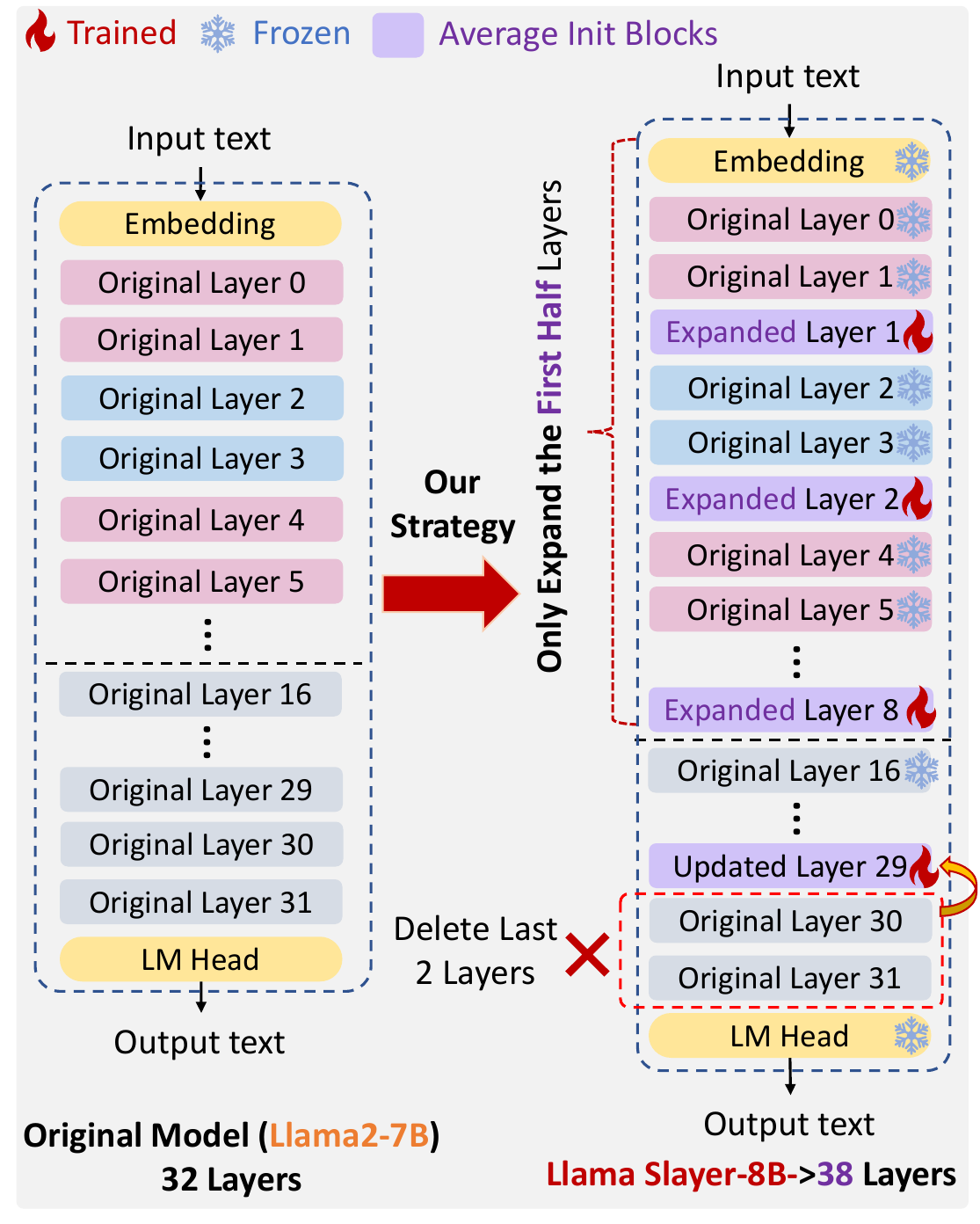}
    \caption{Our strategy focuses on infusing domain-specific knowledge to the first half of the model layers (shallow layers) and the final layer. This is achieved by augmenting the model via block expansion after the deletion of layers deemed non-essential. The expanded blocks are initialized using a linear interpolation technique to ensure a coherent knowledge structure, and only the expanded blocks are fine-tuned.}\label{strategy}
\end{figure}

\subsection{S Strategy}

Based on the above analyses, our S strategy involves expanding the blocks in the areas of the first half (important) of the layer and removing the last few unimportant layers before post-pre-training, as shown in Figure~\ref{strategy}. Only the expanded blocks and final pruned layers get fine-tuned, designated in purple. The block expansion is adopted from Llama Pro \citep{wu2024llama}, but the main differences of our strategy with Llama Pro \citep{wu2024llama} lie in three aspects: (1) we expand blocks at intervals within the shallow (first half) layers based on importance, rather than equally expanding across all layers; (2) We employ a weight averaging method to initialize these expanded blocks, which adds layer coherence and is better than the identity copy used by Llama Pro \citep{wu2024llama}. (3) We remove the less critical final to make the LLM lighter.

Specifically, given an LLM with $L$ blocks, the block expansion incorporates an identity block after each block in the original model, ensuring that the expanded model maintains the same output after expansion. Suppose that we have an initial model with $L$ blocks. Since we regard the shallow layers as more important, we divide the first half ($L/2$) of the original $L$ blocks into $N$ segments, with each segment consisting of $L_N$ blocks. To expand these segments, we replicate the foremost block of each and place it atop the respective segment. The expanded $N$ blocks are then initialized by averaging the weights of two neighboring blocks.  
Considering the limited impact of the deep few LLM blocks, we proceed to delete the last few $D$ layers and the final expanded LLM has $L+N-D$ blocks. To maintain continuity, the final block is reinitialized for training using the combined average weights of the discarded $D$ blocks.

\section{Experiments}

In this section, we detail our key experimental findings. Initially, we discuss our experimental settings (described
in Sec. 4.1), and then provide our post-pretraining results (CPT+SFT) (described in
Sec. 4.2). Finally, ablation
studies of the different block expansion mode, different LLM and data corpus are presented (described in Sec. 4.3).

\begin{table*}
  \centering
  \resizebox{1.99\columnwidth}{!}{
  \begin{tabular}{ccccccc}
    \toprule
    \multirow{2}{*}{Properties} & \multicolumn{2}{c}{CPT datasets (30B token)} & \multicolumn{4}{c}{SFT datasets (1.2M samples)} \\
 \cmidrule(lr){2-3} \cmidrule(lr){4-7}
 & Stack-dedup-Python & Proof-pile-2 & WizardLM evol instruct & SlimOrca & MetaMath & Evol-CodeAlpaca \\
    \midrule
    Total \# Samples & 22B token & 55B token& 0.143M & 0.518M & 0.395M & 0.113M  \\
    Used \# Samples & 8.6B token & 21.4B token & 0.143M & 0.518M & 0.395M & 0.113M  \\
    Open Source & \checkmarkgreen & \checkmarkgreen &\checkmarkgreen & \checkmarkgreen &\checkmarkgreen & \checkmarkgreen  \\
    \bottomrule
  \end{tabular}}
  \caption{Training datasets used for the continued pretraining (CPT) and supervised fine-tuning (SFT) stages, respectively. The ‘Total \# Samples‘ indicates the total number of samples in the entire dataset. The ‘Used \# Samples‘ indicates the actual number of samples that were used in training, which could be no more than the total number of samples in a given dataset. ‘Open Source‘ indicates whether the dataset is open-sourced.}
  \label{tab:dataset_comparison}
\end{table*}

\subsection{Experimental Settings.}

\subsubsection{CPT details.}
We construct a dataset that concentrates on code and math. For the math component, we opt for the Proof-pile-2 dataset\footnote{https://huggingface.co/datasets/EleutherAI/proof-pile-2},
a 55-billion-token amalgamation of scientific papers, web data containing mathematical content, and
mathematical code. As for the code component, the code fragment of our dataset, we draw upon the Stack-dedup dataset\footnote{https://huggingface.co/datasets/bigcode/the-stack-dedup}, a vast repository of openly licensed source codes aggregated from GitHub. Among all the programming languages in Stack-dedup, we only select the 22-billion-token Python division. Notably, due to constraints on our computational resources, we only sample a smaller subset totaling 30 billion tokens from both datasets, maintaining a Math to Code ratio of 5:2, for continued pretraining. In the ablation study section, we further downsize our dataset selection to a 5-billion-token extract, equally distributed following a Math:Code ratio of 5:2, again, to improve experiment efficiency.

Our main experiments are conducted on Llama2-7B. Specifically, we expand the block number from 32 to 38 using an
interleaved approach. In the block expansion process, we only expand the first half 16 layers, setting the parameters to $L_N = 2$ and $N = 8$. This configuration leads to the formation of 8 groups, with each group expanding from 2 blocks to 3 blocks. For the code and math corpus pretraining, we adopt a batch size of 1024 and a sequence length of 4096, combined with a 2$\%$ warmup ratio. The learning rate is set at 2e-4, utilizing a Cosine learning rate decay strategy. To enhance efficiency, we use bf16 mixed precision training, apply a weight decay factor of 0.1, and institute a gradient clipping threshold of 1.0, and apply the flash-attention mechanism during training. Our 30B token experiments were conducted on 256 NVIDIA A100 GPUs to save the training time and our 5B token ablation experiments were conducted on 32 NVIDIA A100 GPUs, all trained for 2 epochs.

% \begin{table}
% \caption{Instruction datasets investigated in this work. We report the average number of rounds ($N_r$), average length of prompts ($L_p$), average length of completion ($L_c$).} \label{sft}
% \centering
% \resizebox{0.99\columnwidth}{!}{
%     \begin{tabular}{ccccccc}
%     \toprule
%         Datasets &Query Source &Response Source & Instances &$N_r$ &$L_p$ &$L_c$\\
%         %\cline{2-9}
%         \midrule
%         %\hline
        
%         WizardLM evol instruct&GPT-4 &GPT-4 &143,000 &1.0 &602.6 &1704.9\\
%         %ShareGPT &User prompts &GPT-3.5/GPT-4 &63,817 &2.9 &293.2 &1157.1\\
%         SlimOrca &Human-written& GPT-4 &517,982 &1.0 &574.3 &599.3\\
%         MetaMath &Human-written/GPT-4 &GPT-4 &395,000 &1.0 &209.4 &498.2\\
%         Evol-CodeAlpaca &GPT-4 &GPT-4 &111,272 &1.0 &652.5 &1552.0\\
%         \bottomrule
%     \end{tabular}}
% \end{table}

\subsubsection{SFT details}

% During the instruction fine-tuning phase, we follow \cite{\cite{wu2024llama}} and amalgamate four distinct data sources to forge the final SFT dataset of our Llama Slayer-8B-Instruct. Specifically, we incorporate the WizardLM evolution instruction dataset \cite{xu2023wizardlm}, which teems with instructive entries of various complexity levels. We also integrate the evolution CodeAlpaca dataset \cite{luo2023wizardcoder}, which consists of ChatGPT-generated complex coding questions alongside the corresponding solutions. Additionally, we use MetaMath \cite{yu2023metamath}, which reframes questions from multiple perspectives, and SlimOrca \cite{slimorca}, a curated subset of our OpenOrca data. SlimOrca provides an efficient route to achieve performance comparable to using larger data slices, while only incorporating approximately 500,000 GPT-4 completions.

During the instruction fine-tuning phase, we follow \cite{wu2024llama} and amalgamate four distinct data sources (WizardLM evolution instruction dataset \cite{xu2023wizardlm}, evolution CodeAlpaca dataset \cite{luo2023wizardcoder}, MetaMath \cite{yu2023metamath} and SlimOrca \cite{slimorca}) to forge the final SFT dataset of our Llama Slayer-8B-Instruct. The final SFT dataset comprises upwards of 1.2M samples. To fine-tune the basic models, we employ
specific configurations, including a batch size of 128, a sequence length of 4096, 0.02 warmup ratio, a
learning rate of 1e-5, a Cosine learning rate scheduler, and bf16 mixed precision.

\subsubsection{Evaluation Metrics} We conduct a comparative analysis of Llama2-7B expanded with our proposed S strategy with the latest state-of-the-art
(SOTA) LLMs. We adopt seven datasets as benchmarks for evaluation: \textit{ARC} (25-shot)\cite{clark2018think}, \textit{HellaSWAG} (10-shot)\cite{zellers2019hellaswag},
\textit{MMLU} (5-shot)\cite{hendrycks2020measuring}, \textit{TruthfulQA} (0-shot) \cite{lin2021truthfulqa}, \textit{Winogrande} (5-shot) \cite{sakaguchi2021winogrande}, \textit{GSM8K} (5-shot) \cite{cobbe2021training} and \textit{HumanEval} (0-shot)\cite{chen2021evaluating}. Also, the average scores for the seven tasks are given. Among these benchmarks, the first five are employed to test the basic knowledge abilities, and GSM8K \citep{cobbe2021training} and HumanEval \citep{chen2021evaluating} are used to test math and coding abilities, respectively. We employ the BigCode Evaluation Harness\footnote{https://github.com/bigcode-project/bigcode-evaluation-harness} to evaluate HumanEval and the Eleuther AI Language Model
Evaluation Harness\footnote{https://github.com/EleutherAI/lm-evaluation-harness} to evaluate the other six benchmarks.

\begin{table*}

\centering
\resizebox{1.99\columnwidth}{!}{
    \begin{tabular}{cccccc|cc|c}
    %\toprule
    \toprule[1pt]
        \multirow{2}{*}{\textbf{Model}} &\multicolumn{5}{c}{\textbf{Language Tasks}} & \multicolumn{2}{c}{\textbf{Math $\&$ Code Tasks}} & \multirow{2}{*}{\textbf{Avg.}}\\
        &ARC &  HellaSwag &MMLU &TruthfulQA & Winogrande & GSM8K &HumanEval\\
        \midrule
        %\hline
        \textit{Pretrained Comparison}\\
        \rowcolor{gray!30}
        LLaMA2-7B \cite{touvron2023llama2} $\diamondsuit$&54.18 &78.59 &45.87 &38.76 &74.03 & 13.12& 12.83& 45.34\\
        LLaMA-7B \cite{touvron2023llama}&50.94 &77.81 &35.69 &34.33 &71.43  & 8.04& 10.61& 41.26\\
        CrystalCoder (7B) \cite{liu2023llm360}&47.01 &71.97 &48.78 &35.91 &67.17 &10.77 & 28.38& 44.28\\
        CodeLLaMA-7B \cite{roziere2023code}&39.93 &60.80 &31.12 &37.82 &64.01 &5.16& 33.50& 38.91\\
        StarCoder-15B \cite{li2023starcoder}&30.38 &47.93 &29.96 &41.28 &56.12 &9.48& 15.32& 32.92\\
        OpenLLaMA-v2-7B \cite{geng2023openllama}&43.69 &72.20 &41.29 &35.54 &69.38 &3.49& 15.32& 40.13\\
        Falcon-7B \cite{almazrouei2023falcon}&47.87 & 78.13 &27.79 &34.26 &72.38 & 4.62& 9.62& 39.24\\
        Llama Pro (8.3B) $\clubsuit$  \cite{wu2024llama} &53.67 & 77.83& 47.17& 37.37& 72.30& 17.02 \textcolor{red}{\small (+3.9$\%$)}& 20.95 \textcolor{red}{\small (+8.1$\%$)}& 46.62\textcolor{red}{\small (+1.3$\%$)}\\
        %Llama Pro (8.3B) \cite{wu2024llama} &54.10 &77.94 &47.88 &39.04 &73.95 &17.89 & 28.66& 48.49\\
        \rowcolor{blue!20}
        Llama SLayer-8B & 54.35 & 77.46  & 48.25 & 36.13& 73.88 &  \textbf{19.03}  \textcolor{red}{\small (+5.9$\%$)}& \textbf{22.62} \textcolor{red}{\small (+9.8$\%$)} &\textbf{47.39}\textcolor{red}{\small (+2.1$\%$)}\\
        %Llama-SL (8B) & 53.16 &   & 44.95 & 37.28& 73.48 &  16.09  \textcolor{red}{\small (+3.0\%)}& 19.52 \textcolor{red}{\small (+6.7\%)} &\\
        \midrule
        
        \textit{SFT Comparison}\\
        %LLAMA PRO-INSTRUCT \cite{wu2024llama}&52.30 &76.88 &52.57 &48.80 &72.53 &43.59& 44.51& \\
        %LLaMA2-7B-Chat \cite{touvron2023llama2}&52.90 &78.55 &48.32 &\textbf{45.57} &71.74 &7.35&14.63 &\\
        \rowcolor{gray!30}
        LLaMA2-7B-Chat \cite{touvron2023llama2}&52.90 &78.55 &48.32 & 45.57 &71.74 &23.95&14.63 &47.95\\
        CodeLLaMA-7B-Instruct \cite{roziere2023code}&36.52 &55.44 &34.54 &41.25 &64.56 &7.96& 34.80& 39.30\\
        %WizardMath-7B  & 54.10& 79.55&45.97 &43.65 & 72.69&2.73 & 12.20& 44.41\\
        Mammoth-7B \cite{yuemammoth} & 49.15& 75.72& 42.29& 38.98& 70.88&53.6 & 10.98& 48.94\\
       %Qwen1.5-7B-Chat \cite{bai2023qwen}& 55.89 & 78.56& 61.7& 53.65& 67.8& 13.19& 29.26&51.44\\%old leaderboard version
       
       %Qwen1.5-7B-Chat \cite{bai2023qwen}& 56.22 & 78.57&60.10 & 53.53& 66.69& 55.17& 29.26&57.08\\
       Falcon-7B-Instruct \citep{almazrouei2023falcon}&45.82 & 70.78& 25.66& 44.07& 68.03& 4.70&$-$ &$-$\\
    
       LLAMA PRO-INSTRUCT (8.3B) $\clubsuit$ \cite{wu2024llama} &51.21 & 76.62& 50.12& 41.13& 71.53&50.18\textcolor{red}{\small (+26.2$\%$)}& 32.92\textcolor{red}{\small (+18.3$\%$)}& 53.39\textcolor{red}{\small (+5.4$\%$)}\\
 \rowcolor{purple!20}
        Llama Slayer-8B-INSTRUCT & 48.98 & 76.91 & 50.34& 44.65& 71.12 & \textbf{56.25}\textcolor{red}{\small (+32.3$\%$)} & \textbf{35.15}\textcolor{red}{\small (+20.5$\%$)}& \textbf{54.77}\textcolor{red}{\small (+6.8$\%$)}\\
        %\bottomrule
        \bottomrule[1pt]
    \end{tabular}
    }
\caption{Comparison of evaluation results among several prominent code and language models. The figures in bold mark the highest ones in each column. The Llama Pro (8.3B) $\clubsuit$ means that this is our self-continually pre-trained version on our 30B token training corpus. The red percentage increases in parentheses are relative to Llama2-7B. The evaluation results of other pre-trained and chat models are adopted from the Open LLM Leaderboard.} \label{table1}
\end{table*}

\subsection{CPT $\&$ SFT Results}
We evaluated the performance of our SLayer-8B model against a series of state-of-the-art pretrained models of similar size. This comparison encompassed both general purpose models, such as LLaMA Pro 8.3B \cite{wu2024llama}, LLaMA2-7B \cite{touvron2023llama2}, and Falcon-7B \citep{almazrouei2023falcon}, as well as coding-specialized models, such as CodeLLaMA \cite{roziere2023code} and math-specialized models, such as Mammoth-7B \cite{yuemammoth}. The results are detailed in Table~\ref{table1}.

The results highlight that SLayer-8B effectively balances natural language processing and math and coding
capabilities. Not only retains the general capabilities of its base model, LLaMA2-7B \cite{touvron2023llama2}, more effectively than Llama Pro, it excels in mathematical and coding tasks. In contrast, CodeLLaMA-7B \cite{roziere2023code} opts to compromise its overall performance to improve its coding proficiency. This enhancement is credited to our expansion strategy, which was developed based on empirical research. By directing specific knowledge to the crucial layers, freezing the initial blocks of LLaMA, and training the expansion blocks initialized via interpolation, we achieve more effective knowledge injection while preserving the model's general strengths. SFT often leads to more significant improvements in the evaluation metrics compared to CPT. This is because the evaluation of an LLM assesses its understanding of questions, response standardization, and general knowledge. Although the CPT process helps the model acquire a broad range of knowledge, it may not enhance response standardization as effectively. In contrast, SFT specifically trains the model to follow instructions more accurately and generate more standardized and precise responses, leading to greater improvements in evaluation metrics.

\begin{table*}
\centering
\resizebox{1.99\columnwidth}{!}{
    \begin{tabular}{cccccc|cc|c}
    %\toprule
    \toprule[1pt]
        \multirow{2}{*}{\textbf{Block Expansion Mode (5B token data)}} &\multicolumn{5}{c}{\textbf{Language Tasks}} & \multicolumn{2}{c}{\textbf{Math $\&$ Code Tasks}} & \multirow{2}{*}{\textbf{Avg.}}\\
        &ARC &  HellaSwag &MMLU &TruthfulQA & Winogrande & GSM8K &HumanEval\\
        \midrule
        \rowcolor{gray!30}
        LLaMA2-7B \cite{touvron2023llama2} &54.18 &78.59 &45.87 &38.76 &74.03 & 13.12& 12.83& 45.34\\
         $(4+\textcolor{red}{1})\times8$ (Llama Pro)
         &51.37 &78.12 &44.36 &37.42 &72.69  & 14.73& 18.08& 45.25\\
        $(2+\textcolor{red}{1})\times3|(4+\textcolor{red}{1})\times3|(7+\textcolor{red}{1})\times2$  &52.05 &78.06 &44.88 &37.09 &73.32 & 15.19& 18.23& 45.55\\
        $(2+\textcolor{red}{1})\times3|(7+\textcolor{red}{1})\times2|(4+\textcolor{red}{1})\times3$ &52.13 &77.85 &45.37 &37.07 &73.48  & 14.86&18.02 & 45.54\\
        
        $(4+\textcolor{red}{1})\times3|(2+\textcolor{red}{1})\times3|(7+\textcolor{red}{1})\times2$  &52.47 &78.16 &45.46 &38.32 &73.70 & 15.54&18.69 & 46.05\\
        $(4+\textcolor{red}{1})\times3|(7+\textcolor{red}{1})\times2|(2+\textcolor{red}{1})\times3$ &52.22 &78.02 &44.55 &38.42 &73.01  & 15.39& 18.20& 45.77\\
        
        $(7+\textcolor{red}{1})\times2|(4+\textcolor{red}{1})\times3|(2+\textcolor{red}{1})\times3$  &52.13 &79.23 &45.44 &35.61 &74.74 &14.71 & 17.68& 45.65\\
        $(7+\textcolor{red}{1})\times2|(2+\textcolor{red}{1})\times3|(4+\textcolor{red}{1})\times3$ &51.02 &77.88 &45.66 &36.37 &74.27  & 15.01& 17.59& 45.40\\
        $16|(2+\textcolor{red}{1})\times8$ &52.88 &78.01 &44.54 &37.28&73.48  & 14.72& 16.68& 45.37\\
        $(2+\textcolor{red}{1})\times8|16$  & 53.05&  77.10 & 44.52 & 37.98& 73.11 &  15.96 & 19.50 & 45.89\\
        
        $(2+\textcolor{red}{1})\times8|16  \diamondsuit$ &53.16 & 77.25  & 44.95 & 38.28& 73.48 &  16.09  & 19.52  & 46.10\\

        $(2+\textcolor{red}{1})\times6|17|\textcolor{red}{1}\diamondsuit$ &51.16 & 77.20&45.89 &37.18&72.96&15.52&18.51&45.49\\

        $(2+\textcolor{red}{1})\times8|11|\textcolor{red}{1}\diamondsuit$ (delete 4 blocks, Ours)&49.83 & 75.52& 45.06& 37.33& 73.24 & 13.72& 16.08& 44.40\\
        $(2+\textcolor{red}{1})\times8|12|\textcolor{red}{1}\diamondsuit$ (delete 3 blocks, Ours)&50.15 & 76.02& 45.22& 37.64&  73.30&15.13 &17.96 & 45.06\\
        \rowcolor{blue!20}
        $(2+\textcolor{red}{1})\times8|13|\textcolor{red}{1} \diamondsuit$ (delete 2 blocks, Ours) &51.45 & 77.73  & 46.26 & 37.91& 73.56 &  \textbf{16.37}&  \textbf{19.73} & \textbf{46.14}\\
        %\bottomrule
        \bottomrule[1pt]
    \end{tabular}}
\caption{Comparison of evaluation results among several prominent code and language models. The numbers in red denote that the expanded blocks and only these blocks get fine-tuned at the continued pre-training stage. Methods marked with $\diamondsuit$ indicate that the expanded blocks use weight averaging for initialization. In contrast, other methods use identity copy-for initialization.} \label{ab1}
\end{table*}

\begin{table}
\caption{Comparison of parameters and training time.} \label{table_complex}
\centering
\resizebox{0.99\columnwidth}{!}{
    \begin{tabular}{c|ccc}
    \toprule
        Method & Trainable Parameters (B)&Total Parameters (B)&CPT Time Cost (h)\\
        %\cline{2-9}
        \midrule
        %\hline
        %MDvsFA \cite{wang2019miss}  & 6.28& 1644.7&0.24\\
        Llama Pro & 1.75&8.3&740\\
       Llama Slayer-8B & 1.95 &7.9&700\\
        \bottomrule
    \end{tabular}}
\end{table}

\subsection{Efficiency Comparison}
We have compared in Table~\ref{table_complex} the cost of our Llama Slayer 8.3B and Llama Pro 8B in terms of parameters and training time in our 30B token CPT dataset. The training time is fairly compared to the same settings on 8 NVIDIA A100 GPUs. Although our model's trainable parameters are slightly higher, our model still saves 5.5$\%$ training time cost compared to Llama Pro. In addition, since our model requires less parameters, it is more efficient in storage and inference.

\subsection{Ablation Studies}
%\subsubsection{Ablation study of different block expansion mode.} 
The ablation study of different block expansion mode is shown in Table~\ref{ab1}. In particular, here the data set for ablation is the version of the 5B token extract, which is different from the 30B CPT data set used in Table~\ref{table1} to improve the efficiency of the experiment. To clarify, the notation $(2+\textcolor{red}{1})\times8|16$ means that we expand the model by adding one block for every two blocks of the head eight times, and only the eight expanded blocks marked red will be fine-tuned.

For fair comparison, we first set the trainable parameter number to the same and explored varying densities of block expansion throughout the layers. This strategy aims to identify the most advantageous segments for expansion within the LLM during continued pre-training. Our findings highlight a preference for expanding the shallow layers over a uniform distribution across the entire network. %This observation stems from the diminished effectiveness of deep layers, as revealed in Fig.~\ref{charts}.

We also explored the impact of expanding range by comparing with expanding within the first 1/3 of the layer range, rather than the first 1/2, to reduce computational costs. The results of this approach are presented in Table~\ref{table1}. Specifically, since the range is limited to the first 1/3 of the layers (approximately 12 layers), we expanded 1 layer for every 2 consecutive layers, repeating this process 6 times starting from the first layer. We then removed the last 2 layers and performed a weighted average on the final layer. This method is denoted as $(2+\textcolor{red}{1})\times6|17|\textcolor{red}{1}\diamondsuit$, where the number '1' marked in red indicates that only the red colored layers (the expanded 6 layers and the final 1 layer) get tuned. The result is that the overall performance of expanding within the first 1/3 of the layers is lower compared to expanding within the first 1/2 of the layers. This can be attributed to the observations in Figure~\ref{charts}(a), which shows a significant drop in angular distance from the middle (16th) layer. This suggests that the layers beyond the first 1/3 play a crucial role in maintaining or improving model performance.

\begin{figure}
    \centering \includegraphics[width=0.50\textwidth,height=0.21\textwidth]{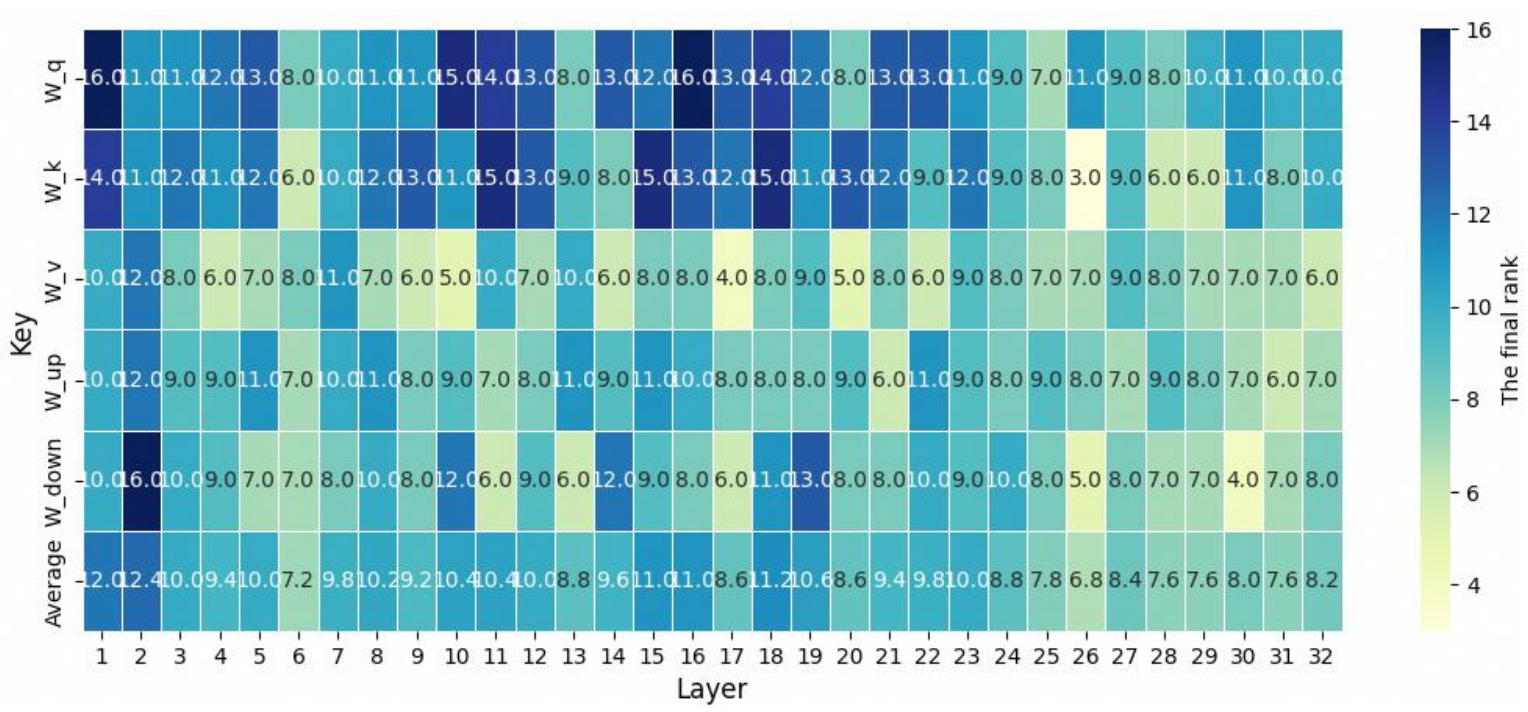}
    \caption{The resulting rank of each incremental matrix when fine-tuning Llama2-7B on our 5B token math+code dataset with AdaLoRA.}\label{heat}
\end{figure}

To further demonstrate the importance of shallow layers, we fine-tune Llama2-7B on our 5B token math code dataset with AdaLoRA \cite{zhang2023adaptive}, which can dynamically allocate parameter budgets to weight matrices based on importance ratings. We display the resulting rank distribution and the average ranks of each incremental matrix during AdaLoRA fintuing in Fig.~\ref{heat}. As shown in Fig.~\ref{heat}, AdaLoRA predominantly enriches the weight matrices in the shallow layers, corroborating our insight that these layers are more crucial for infusing new knowledge. Furthermore, we also calculate the angular distances for Llama Slayer (13.60) and Llama Pro (13.17) after training on our compiled 30B token math+code dataset and observe that the former surpasses the latter, again proving the effectiveness of our approach.

Furthermore, we evaluate the impact of initialization methods and the acceptable limits for layer reduction in Table~\ref{ab1}. We find that weight averaging is a superior method for initializing expanded blocks, since this adds coherence to the expanded LLM. Additionally, we discern that eliminating up to two of the deepest layers is feasible without severely degrading model performance. However, a more aggressive approach to layer deletion has been found to adversely affect the effectiveness of the LLM.

% \begin{table}
% \centering
% \resizebox{0.99\columnwidth}{!}{
%     \begin{tabular}{ccc}
%     \toprule
%         Model & Math & Coding \\
%         %\cline{2-9}
%         \midrule
%         LLaMA2-7B-Chat   & 1.70& 3.10\\
%         LLAMA PRO-INSTRUCT   & 2.60& 4.00\\
%         LLAMA SLayer-8B-INSTRUCT   & 3.11& 4.50\\
%         \bottomrule
%     \end{tabular}}
% \caption{ Llama3-70B automatic evaluation of Chatbot models. LLAMA SLayer-8B-INSTRUCT outperforms widely used LLaMA community chatbots.} \label{table_complex}
% \end{table}

% \begin{figure}
%     \centering \includegraphics[width=0.50\textwidth,height=0.20\textwidth]{latex/curve_after.pdf}
%     \caption{Angular distance between
% inputs and outputs of each block of Llama Slayer and Llama Pro.}\label{curve_after}
% \end{figure}

%\subsubsection{Ablation study of different model$\&$dataset.} 

Apart from the aspect of code and math corpus, we also explore
our training strategy on a different LLM type, Mistral-7B, and another knowledge domain: law, with the
free-law subset of the Pile dataset as our pre-training corpus
\cite{gao2020pile}. To assess our model's proficiency in legal language, we leveraged the Unfair ToS dataset, which is composed of Terms of Service (ToS) agreements from various online platforms—a critical resource in evaluating legal document comprehension. Our evaluation was carried out using the UNFAIR-ToS benchmark \cite{lippi2019claudette} within LexGLUE \cite{chalkidis2021lexglue}, employing a 5-shot learning scenario. The evaluation was also implemented through the Eleuther AI Language Model Evaluation Harness.

\begin{figure}
    \centering \includegraphics[width=0.49\textwidth,height=0.16\textwidth]{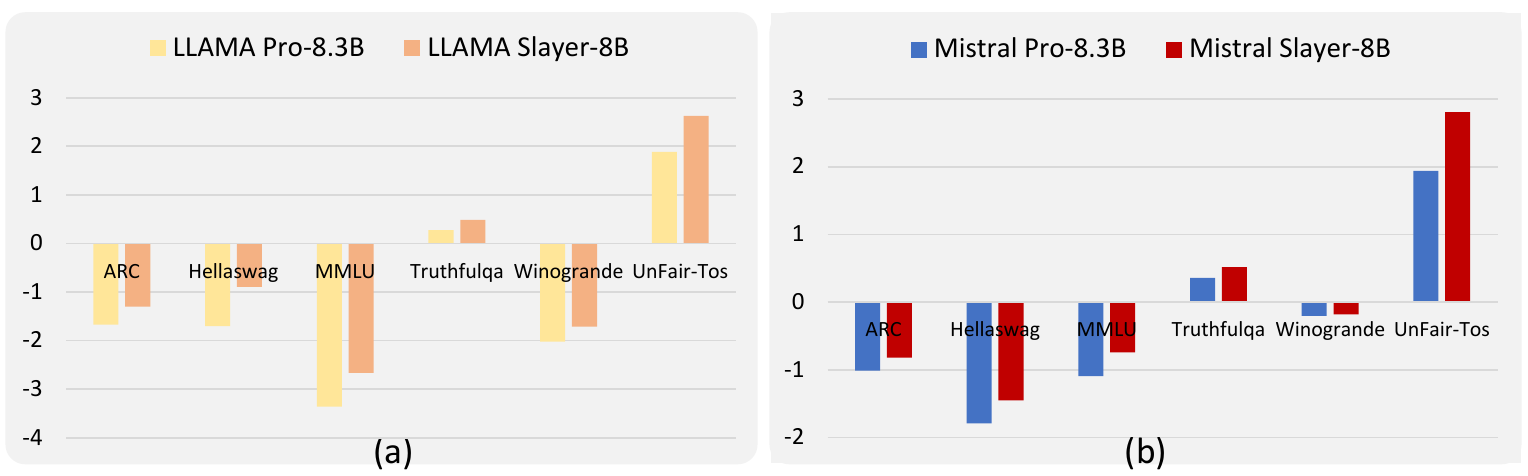}
    \caption{Comparative analysis of performance variations of different training strategies relative to the base model (a) Llama2-7B and (b) Mistral-7B, on both general and law-specific tasks.}\label{law}
\end{figure}

In Fig.~\ref{law}, we detail a comparative analysis of performance variations between our Llama/Mistral Slayer-8B models and the self-post pretrained Llama/Mistral Pro, benchmarked against the corresponding foundational models. It is evident from our findings that our tailored strategy significantly enhances domain-specific knowledge injection while more effectively mitigating the issue of catastrophic forgetting compared to the Llama Pro's even block expansion. This improvement can be largely attributed to our idea of expanding LLM blocks based on their layer importance. Such an approach ensures that our knowledge augmentation efforts are concentrated on the most crucial blocks, thereby enhancing the efficiency of post-pretraining. In addition, our utilization of the weight-averaging technique for initializing the expanded blocks provides a smoother approach compared to the direct identity copy employed in Llama Pro. This ensures better layer coherence in the model, as opposed to the abrupt identity copy method. Therefore, our proposed strategy not only fosters smoother integration and adaptation of new knowledge, but also contributes to the refined performance and learning capabilities of our models within domain-specific contexts.

% \begin{table*}
% \caption{Comparison of evaluation results among different training strategies, reporting performance on both general and law-specific tasks.} \label{ab2}
% \centering
% \resizebox{1.99\columnwidth}{!}{
%     \begin{tabular}{cccccc|c|c}
%     %\toprule
%     \toprule[1pt]
%         \multirow{2}{*}{\textbf{Model}} &\multicolumn{5}{c}{\textbf{Language Tasks}} & \multicolumn{1}{c}{\textbf{Law Task}} & \multirow{2}{*}{\textbf{Avg.}}\\
%         &ARC &  HellaSwag &MMLU &TruthfulQA & Winogrande & Unfair-Tos &\\
%         \midrule
% Mistral-7B  & 61.43& 82.99& 62.64& 42.62& 79.24& 84.25&  \\
% Mistral-7B-SL & 47.61& 72.54& 30.90& 43.14&69.06 &87.06 &  \\
% Mistral-7B-AD  & & & & & & 86.19&  \\
% \midrule
% LLAMA2-7B  &54.18 &78.59 &45.87 &38.76 &74.03 &88.48&  \\
% LLAMA2-7B-SL &51.88 & 77.70& 41.11& 39.25& 71.12& 90.01& \\
% LLAMA2-7B-AD &52.13 & 77.54&41.88 & 40.03& 71.03& 89.58&  \\
%         %\bottomrule
%         \bottomrule[1pt]
%     \end{tabular}
%    } 
    
% \end{table*}  

% \begin{figure*}[t]
%   \includegraphics[width=\columnwidth]{example-image-golden}
%   \caption{A figure with a caption that runs for more than one line.
%     Example image is usually available through the \texttt{mwe} package
%     without even mentioning it in the preamble.}
%   \label{fig:experiments}
% \end{figure*}

% \begin{figure*}[t]
%   \includegraphics[width=0.48\linewidth]{example-image-a} \hfill
%   \includegraphics[width=0.48\linewidth]{example-image-b}
%   \caption {A minimal working example to demonstrate how to place
%     two images side-by-side.}
% \end{figure*}

\section{Conclusion}

We propose S strategy to inject domain-specific knowledge into LLMs via block expansion based on layer importance in the post-pretraining phase. This strategy prioritizes knowledge injection to the important shallow layers while pruning the ineffective deep layers, and can not only bolster the model's specific-domain capabilities but also maintain its general proficiency. Based on the proposed strategy, we introduce LLAMA SLayer-8B and LLAMA SLayer-8B-INSTRUCT, LLMs that derive from the base model LLaMA2-7B. The two models surpass various predecessors in the LLaMA series across a wide array of benchmarks, evidencing the superior performance of our strategy.

\section{Limitations}

% Although our study presents a promising method
% for balancing general and domain-specific capabilities in LLMs, its scope is limited to the language modality, especially programming language and English. Future research could explore extending
% the application of our block expansion method to
% other domains, such as maintaining original language ability in multimodal large language models(Ge et al., 2023; Bai et al., 2023), and multilingual domains. Also, whether the proposed strategy fits larger LLMs and whether we can delete more deep layers when the LLM get larger remains to be explored in the future.

While we introduces an effective strategy based on layer importance to post-pretrain Large Language Models (LLMs) to achieve a balance between general and domain-specific abilities, its applicability is primarily within the realm of language. Future investigations could aim to broaden the utility of our knowledge injection strategy across various fields. This expansion might include adapting the technique to enhance the native language capabilities of multimodal LLMs \cite{ge2023making}, as well as its application in multilingual settings. Furthermore, the adaptability of our proposed strategy to larger LLMs and the feasibility of removing an increased number of deeper layers as the model size escalates are promising avenues for future research.

% \subsection{Appendices}

% Use \verb|\appendix| before any appendix section to switch the section numbering over to letters. See Appendix~\ref{sec:appendix} for an example.

%\section*{Acknowledgments}

\bibliography{custom}

\appendix

\section{Evaluation Benchmark}
\label{sec:appendix}
The benchmarks used for evaluation include:

\begin{itemize}
    \item AI2 Reasoning Challenge \cite{clark2018think}(25-shot): a set of grade-school science questions.
    \item HellaSwag (10-shot) \cite{zellers2019hellaswag}: a test of commonsense inference, which is easy for
humans (approximately 95$\%$) but challenging for SOTA models.
    \item TruthfulQA (0-shot) \cite{lin2021truthfulqa}: a test to measure a model’s propensity to reproduce falsehoods commonly found online.
    \item MMLU (5-shot) \cite{hendrycks2020measuring}: a test to measure a text model’s multitask accuracy. The
test covers 57 tasks including elementary mathematics, US history, computer science, law, and more.
    \item Winogrande (5-shot) \cite{sakaguchi2021winogrande}: an adversarial and difficult Winograd benchmark at
scale, for commonsense reasoning.
    \item GSM8k (5-shot) \cite{cobbe2021training}: diverse grade school math word problems to measure a model’s
ability to solve multi-step mathematical reasoning problems.
    \item HumanEval (0-shot) \cite{chen2021evaluating}: 164 handwritten Python programming problems with a
function signature, docstring, body, and several unit tests.
\end{itemize}

\section{Hyper-parameters of pretraining on the domain of law.}

We list the detailed settings of our pretraining on the domain of law in Table~\ref{law_setting}.
\begin{table}
\caption{ Hyper-parameters of pretraining on the domain of law.} \label{law_setting}
\vskip 0.15in
\begin{center}
\begin{small}
\begin{sc}
\centering
\resizebox{0.99\columnwidth}{!}{
    \begin{tabular}{cc}
    \toprule
        Hyper-parameters & Assignment\\
        %\cline{2-9}
        \hline
        %\hline
        Batch size &1024\\
Maximum sequence length &2,048\\
Maximum learning rate &2e-4\\
Optimizer &Adam\\
Adam beta weights &0.9, 0.95\\
Learning rate scheduler &cosine\\
Warmup ratio &0.02\\
Gradient clipping &1.0\\
Epoch &2\\
        \bottomrule
    \end{tabular}}\label{setting}
\end{sc}
\end{small}
\end{center}
\vskip -0.1in
\end{table}

\section{More detailed explanation of Figure 4.}

Figure 4 shows the resulting rank of each incremenal matrix when fine-tuning Llama2-7B
on our 5B toke nmath+dode data with AdaLoRA. The y axis has six weight items, $W_q,W_k,W_v,W_{up},W_{down}$ and the average weight value. From the last average line we can see that AdaLoRA allocates more parameter budget to the weight
matrices in shallow layers, since the color gradually becomes lighter as the layer gets deeper. Such behavior aligns with our conclusion that the shallow layers are more important to knowledge injection and should be paid more attention during post-pretraining.

\end{document}